\documentclass{article}
\usepackage{spconf}
\usepackage{makecell}
\usepackage{amsmath,graphicx,caption,subcaption,tabularx,float}
\usepackage{booktabs}
\usepackage{xcolor}

\title{Understanding Shared Speech-Text Representations}
\name{\makecell{Gary Wang$^\star$, Kyle Kastner$^\star$, Ankur Bapna, Zhehuai Chen\\\textit{Andrew Rosenberg, Bhuvana Ramabhadran, Yu Zhang} \thanks{$\star$ Equal Contributions}}}
\address{Google LLC, USA}
\begin{document}
\ninept
\maketitle
\begin{abstract}
Recently, a number of approaches to train speech models by incorporating text into end-to-end models have been developed, with Maestro advancing state-of-the-art automatic speech recognition (ASR) and Speech Translation (ST) performance. In this paper, we expand our understanding of the resulting shared speech-text representations with two types of analyses.  First we examine the limits of speech-free domain adaptation, finding that a corpus-specific duration model for speech-text alignment is the most important component for learning a shared speech-text representation.  Second, we inspect the similarities between activations of unimodal (speech or text) encoders as compared to the activations of a shared encoder.  We find that the shared encoder learns a more compact and overlapping speech-text representation than the uni-modal encoders. We hypothesize that this partially explains the effectiveness of the Maestro shared speech-text representations.

\end{abstract}
\vspace{-.3cm}
\section{Introduction}
\label{sec:intro}

The availability of vast amounts of untranscribed speech and text resources has contributed to increased interest in semi-supervised and unsupervised learning approaches for Automated Speech Recognition (ASR). A range of techniques that incorporate text resources at various entry points in an end-to-end ASR model have been explored in the literature. Text injection approaches that either inject text into the encoder~\cite{bapna2021slam, bapna2022mslam, chen2022maestro, sainath2022joist} or decoder~\cite{meng2022mhat,meng2022ilma} have shown success in low resource applications as well as domain and language transfer~\cite{thomas2022textogram, chen2022maestro, sainath2022joist}, while rescoring approaches~\cite{udagawa2022rescoring} use first-pass recognition hypotheses to rescore with a large language model.

Maestro \cite{chen2022maestro} is an approach to train a speech model using a joint speech-text representation on transcribed speech, untranscribed speech and unspoken text.  The joint speech-text representation is learned in two ways. First, using the paired data, activations from two modal encoders, the speech encoder and the text encoder, are aligned and optimized to be similar through a consistency loss term.  Second, the speech encoder activations and text encoder activations are passed through a common shared encoder. The output of the shared encoder is the joint speech-text representation that serves as input to the ASR decoder (RNN-T ~\cite{graves2012sequence,graves2013speech}). These jointly learned representations have yielded state-of-the-art results in not only well-benchmarked, monolingual and multilingual ASR as well as speech translation but have further demonstrated the richness of these learned representations with the ability to build ASR systems for languages with no transcribed speech~\cite{chen2022maestro-u}. 

While we know that joint speech-text representation learning improves ASR, our understanding of the value and structure of the learned representations is less well developed. 

In this work, we expand on this understanding in two directions. %using Maestro as proposed in~\cite{chen2022maestro}.  
First, we evaluate the ability to transfer information from one domain to another through the joint representation (Section \ref{sec:domain-adaptation}).   
We explore which components of the text encoder are robust across corpora, and which are sensitive.
Second, we investigate the modal representations from the speech and text encoders (Section \ref{sec:analysis}).  We inspect the cross-modal consistency loss as a signal of robustness, and the ability for this loss term to generalize across corpora through T-SNE visualization of activations and a retrieval probe task.  

In this analysis, we compare the representations learned by Maestro \cite{chen2022maestro} and SLAM \cite{bapna2021slam}. The contributions of this work are:

\vspace{.2cm}
\textbf {Speech-Free Domain Adaptation (Section \ref{sec:domain-adaptation})}
    
\begin{itemize}

    \item We show that for speech-free domain adaptation with speech-text representations, duration/alignment is the most important aspect to model. Corpus specific text-encoders are not substantially better than using a general purpose encoder with a corpus specific duration model.
    \item We demonstrate that Maestro enables speech-free domain adaptation with only text data using the corpora in SpeechStew~\cite{chan2021speechstew}. 
    \end{itemize}

\textbf{ Representation Space Analysis (Section~\ref{sec:analysis})}
    \begin{itemize}
    \item We show that the Maestro shared encoder learns a unified shared speech-text representation space. However, the modal encoders, even when trained with a consistency loss, learn distinct representations . 
    \item Using a cross-modal retrieval task as a probe, we demonstrate that the shared encoder representations provide better retrieval performance.
\end{itemize}

\section{Related Work}

The shared speech-text representations we explore in this paper are based on an architecture in which data is passed through modal encoders, whose activations are then consumed by a shared encoder.  This structure is used by SLAM~\cite{bapna2021slam} and mSLAM~\cite{bapna2022mslam} where the speech and text encoder outputs are concatenated. Maestro extended this to align the modal activations and present them independently to the shared encoder.  Related approaches either eliminate this alignment step entirely, using a fixed upsampling of text embeddings~\cite{thomas2022textogram} or use a simpler, deterministic or independent Gaussian duration model~\cite{sainath2022joist}.  In contrast to an independent text encoder, \cite{sato2022textonly} performs aligned-text injection using residual adapters in the speech encoder.  Alternately, SpeechBert~\cite{chuang2019speechbert}
 and~\cite{sunder2022tokenwise} operate by unifying speech and language model embeddings.

Canonical Correlation Analysis (CCA) between different modalities is a classical technique \cite{hotelling1936relations} that enables a more direct approach to merging representations.  Later CCA was extended to DeepCCA, drawing from the original formulations, but incorporating neural networks e.g \cite{sun2020dcca}. Acoustic word embeddings take similar inspiration from CCA, learning joint projection spaces e.g.~\cite{hu2020multilingual, kamper2016acousticword}.

\section{Architecture and Training}
\label{sec:architecture-and-training}

\begin{table*}[!h]
  \caption{LibriSpeech Maestro adaptation results.}
  \label{t:topline_text_adaptation_results}
  %\vskip 0.1in
  \centering
  \small
  \resizebox{1.60\columnwidth}{!}{%
  \begin{tabular}{lccccccccc}
    \toprule
    \bfseries Method & \bfseries Pair Data & \bfseries Text Data &
    \multicolumn{2}{c}{LibriSpeech} & \multicolumn{2}{c}{AMI} &
    \multicolumn{1}{c}{CV} & \multicolumn{1}{c}{SWBD} & \multicolumn{1}{c}{TED} \\
    \cmidrule{4-10}
    &&& test-clean & test-other & ihm & sdm1 & test & test & test\\
    \midrule
    \bfseries LS Maestro init & LS &LS LM&  2.22 & 4.27 & 35.42 & 60.02 & 22.48 & 31.77 & 7.73\\
    \midrule
    \bfseries Text Adaptation \\
    \quad AMI & LS & AMI Sup & 1.57 & 3.09 & \bfseries 30.26 & \bfseries 54.52 & 22.46 & 29.90 & 7.10 \\
    \quad CV & LS & CV Sup & 1.57 & 3.06 &34.21 & 57.13 & 21.16 & 31.19 & 7.11 \\
    \quad SWBD & LS & SWBD Sup & 1.51 & 3.07 & 31.19 & 55.40 & \bfseries 20.70 & \bfseries 28.97 & 6.66 \\
    \quad TED & LS & TED Sup + LM & \bfseries 1.45 & \bfseries 3.05 &  32.08 & 55.36 & 21.03 & 29.83 & \bfseries 6.05 \\
    \bottomrule
  \end{tabular}
  }
  \vspace{-.15cm}
\end{table*}

\subsection{Maestro Architecture}
\label{sec:architecture-and-training:asr-maestro-architecture}

\begin{figure}[H] 
\centering
\includegraphics[width=.9\columnwidth]{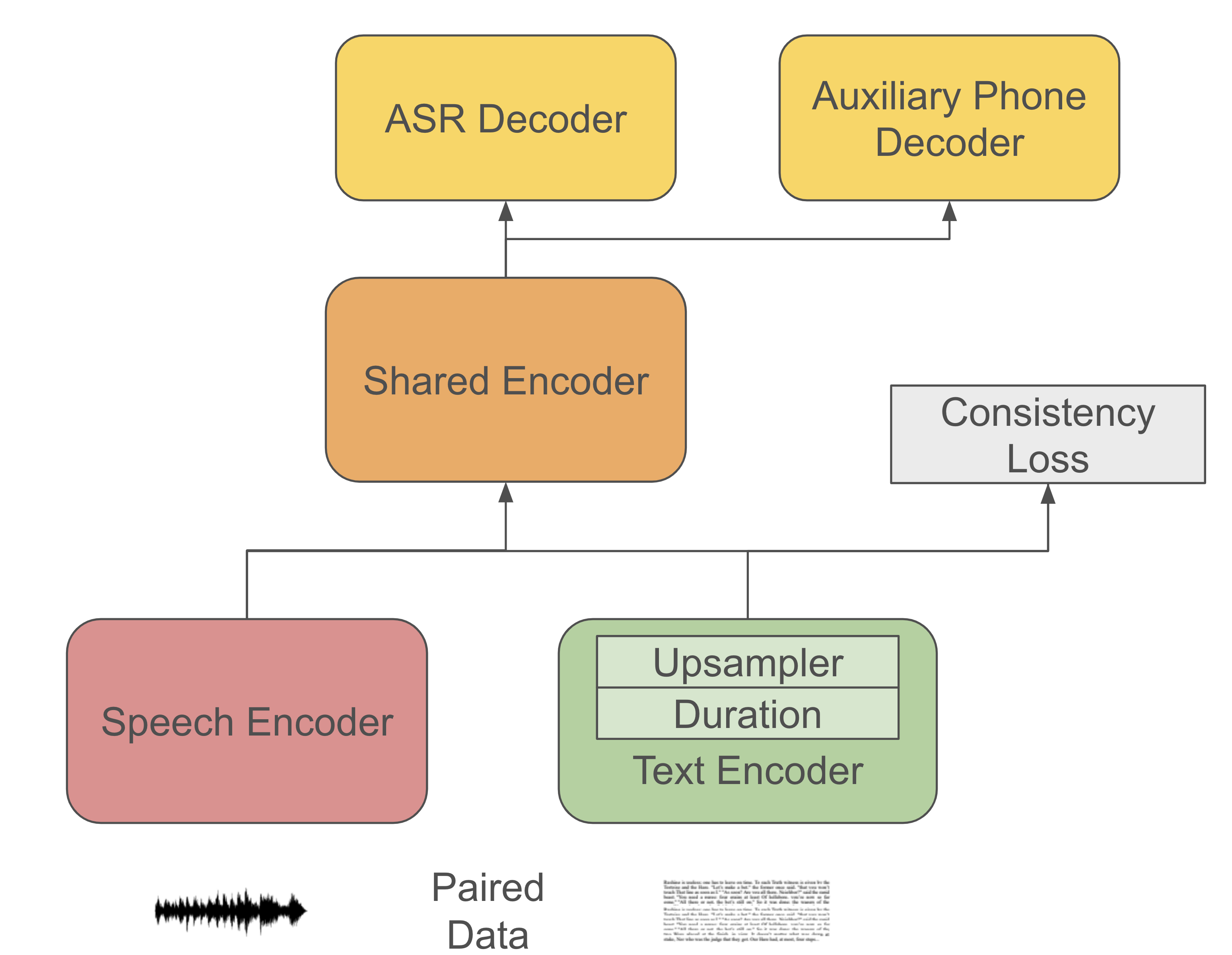}
\caption{Maestro architecture.}
\label{fig:maestro_architecture}
\end{figure}

The Maestro architecture~\cite{chen2022maestro} consists of a $600$ million parameter full context Conformer transducer that splits a standard ASR encoder into two sub-encoders, a speech encoder, and a shared encoder and introduces a uni-modal text encoder (See Figure \ref{fig:maestro_architecture}).  
The speech encoder ingests log-mel features as input and the shared encoder takes the output of the speech or text modal encoders. The speech encoder consists of the first $6$ conformer layers, and the shared encoder consists of the last $18$ conformer layers of the base ASR encoder. We use the text encoder architecture described in \cite{chen2022maestro}. The text encoder is trained on paired data, since ground-truth text/wordpiece(wpm) and feature mappings are available. It includes a text/phoneme encoder, a trained duration model and re-sampling layer, and a refiner. During inference, the duration model is responsible for predicting the duration of each input text/phoneme token, and the re-sampling layer upsamples the text representations according to their respective durations. The refiner takes the upsampled representations and refines them further for modality matching. Maestro adds an additional RNN-T phoneme decoder of the same parameter size to train the duration model. 
In contrast to~\cite{chen2022maestro}, unless explicitly mentioned our experiments do not include a second pass fine-tuning.  

\subsection{SLAM Architecture}
\label{sec:architecture-and-training:slam-architecture}

SLAM \cite{bapna2021slam, bapna2022mslam} uses a similar structure to Maestro, decomposing its encoder into a speech encoder, text encoder and shared encoder. The major distinctions are 1) the structure of the text encoder, and 2) training of the model.  First, the SLAM text encoder consists of a text embedding layer, and sinusoidal position embedding.  Second, when training on transcribed data, the outputs of the speech and text encoders are concatenated and used as input to the shared encoder (as opposed to Maestro which consumes {\em either} the speech or text activations).  To align representations and encourage consistency between speech and text, SLAM uses a contrastive Speech-Text Matching loss, and a Translation Language Model loss similar to~\cite{zheng2021fused}. mSLAM~\cite{bapna2022mslam} replaces the Speech-Text matching loss with a CTC loss on the speech representations of the transcribed data.

\section{Text-Only Domain Adaptation}
\label{sec:domain-adaptation}
\vspace{-.2cm}
\subsection{Data}
\label{sec:architecture-and-training:datasets-and-setup}
We use subsets of SpeechStew~\cite{chan2021speechstew} comprising of LibriSpeech, AMI, Common Voice (CV), Switchboard (SWBD), and TED corpora for text adaptation experiments. For all experiments, we initialize from a converged  LibriSpeech (LS) Maestro model 
trained on 3 types of data: LibrilLight~\cite{kahn2020libri} as untranscribed speech, LibriSpeech ($960$ hours) as paired data, and LibriSpeech LM text as unspoken text. For text adaptation, we continue to train the LS Maestro model with text from different SpeechStew corpora.  During adaptation, we construct a minibatch with $256$ text utterances from the target domain, and $256$ transcribed speech utterances from LibriSpeech.  The use of supervised data during text-only adaptation stabilizes training and minimizes catastrophic forgetting.

\subsection{Experiments and Results}
\label{sec:domain-adaptation:topline-results}
We present results of adapting with text only data from various corpora. The text encoder is always trained on LibriSpeech data. Table \ref{t:topline_text_adaptation_results} shows the results of text-only adaptation across $4$ datasets, compared to the LS Maestro baseline.  We can observe that, for all datasets, using \textbf{in-domain text} successfully enables adaptation to the target domain despite using durations predicted on Librispeech, delivering a relative WER improvement as high as \textbf{$20$\%} on TED. We also observe that this speech-free text adaptation benefits other testsets, even though their in-domain text is never observed during training. This observation is the strongest in Switchboard, where Switchboard text adaptation improves performance across AMI, Common Voice and TED testsets with over \textbf{$7+$\%} relative WER gains.

\subsection{Text Encoder Data Quality Ablation}
\label{sec:domain-adaptation:text-encoder-ablation}
Next, we study the impact of domain on text injection performance. The text encoders can be trained on any paired data and the duration model used for upsampling during text injection can be domain specific. As an ablation study on text adaptation, we use the LS Maestro model and corpus specific text encoders trained on the remaining SpeechStew corpora. We study the impact of duration models trained on different domains while using text from the Switchboard (SWBD) corpus alone for adaptation.
We initialize a Maestro model from w2v-BERT~\cite{chung2021w2v} pretrained speech encoder and randomly initialize the text encoder. We subsequently train the text encoder with corpus-specfic paired data for $80$k steps to convergence.
We use the training recipe outlined in section ~\ref{sec:domain-adaptation:topline-results}, where we load and freeze the text encoder during text injection.

Table \ref{t:swbd_text_encoder_ablation} presents the performance measured as WER and MSE consistency loss (text encoder loss) of the different text encoders for text adaptation on SWB.
We hypothesize that the MSE values provide an indication of quality of the text encoders. Supporting this hypothesis, we observe a trend between the Text Encoder Loss and the SWBD text adaptation performance. It is interesting to note that durations trained on the TED corpus yield the best performance on SWB.  This may suggest that TED duration model has higher quality across domains (as indicated by the Text Encoder Loss), or if there is a similarity between SWBD and TED speaking styles.  

\begin{table}[H]
  \captionsetup{justification=centering}
  \caption{Ablation of text encoder trained on different corpora with Switchboard (SWBD) Text Adaptation, using text encoders trained on different corpora.}
  \label{t:swbd_text_encoder_ablation}
  %\vskip 0.11in
  \centering
  \small
  \resizebox{1.00\columnwidth}{!}{%
  \begin{tabular}{lccc}
    \toprule
    \bfseries Method & \bfseries Text Encoder Loss & \multicolumn{1}{c}{SWBD} \\
    \midrule
    \bfseries SWBD Supervised FT (Lower Bound)  & - & 23.24 \\
    \bfseries LS Maestro init & - & 31.77 \\
    \midrule
    AMI Duration Training  & 6.5 & 29.26 \\
    CV Duration Training  & 5.3 & 29.18  \\
    TED Duration Training  & 4.1 & \bfseries 28.45 \\
    \bottomrule
  \end{tabular}
  }
  \vspace{-.15cm}
\end{table}

\subsection{Text Encoder Component Ablation}
\label{sec:domain-adaptation:text-encoder-component-ablation}
We next evaluate the impact of different components of the text-encoder using the AMI corpus for text adaptation. AMI comprises spontaneous elicited meeting speech, while LibriSpeech is read books.
We examine the two main components of the text encoder, the duration model, and the refiner.  The duration model includes a text/phoneme projection, duration prediction and up-sampling layers. The refiner is responsible for taking the upsampled encoding, and refining it to match the speech encoder output. We start with an entirely in-domain AMI text encoder, and then measure the effect of using the AMI-trained duration model, but keeping the LibriSpeech refiner. Note that since the paired training data is always LibriSpeech, using the LibriSpeech refiner allows us to investigate the value of the duration model without introducing the additional complexity of mismatched data used in training the speech encoder and text encoder refiner. 
Table \ref{t:text_encoder_ablation} includes the results.
\begin{table}[H]
  \captionsetup{justification=centering}
  \caption{Ablation of importance for in-domain text encoder components with AMI Text adaptation.}
  \label{t:text_encoder_ablation}
  %\vskip 0.11in
  \centering
  \small
  \resizebox{.9\columnwidth}{!}{%
  \begin{tabular}{lcccc}
    \toprule
    \bfseries Method & \multicolumn{2}{c}{AMI} \\
    \cmidrule{4-5}
    & ihm & sdm1\\
    \midrule
    \bfseries AMI Supervised FT (Lower Bound) & 11.60 & 28.03\\
    \bfseries LS Maestro init & 35.42 & 60.02\\
    \midrule
    LS Duration \& Refiner & 30.26 & 54.52 \\
    AMI Duration \& Refiner &28.78 & 53.82 \\
    AMI Duration, LS Refiner & \bfseries 27.20 & \bfseries 52.13 \\
    \bottomrule
  \end{tabular}
  }
  \vspace{-.15cm}
\end{table}
As we switch to an AMI text encoder entirely, we see an improvement in adaptation performance due to the in-domain text encoder. However, if we use and freeze only the AMI {\em duration} model, and utilize a LibriSpeech refiner, the performance improves even further.  This suggests that the primary value of a domain-specific text-encoder for text-only adaptation comes from having a high quality duration model for the new domain.

It remains to be seen to what degree duration models can be transferred between related corpora. For example, AMI is spontaneous, elicited speech, while LibriSpeech is read.  The impact of the duration model may be less pronounced when adapting from LibriSpeech to another read corpus.

\vspace{-.2cm}

\section{Representation Space Analysis}
\label{sec:analysis}

\begin{figure}[!htb]
    \vspace{-.4cm}
    \centering
    \begin{minipage}{\columnwidth}
        \centering
        \includegraphics[width=.95\textwidth, height=5cm]{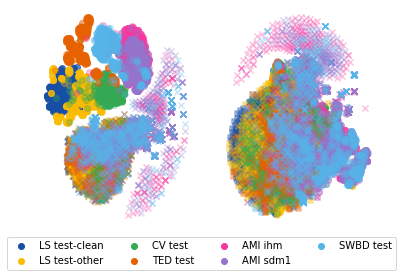}
    \end{minipage}%
    \vspace{-.25cm}
    \captionsetup{justification=centering}
    \caption{T-SNE of LibriSpeech Maestro text (crosses) and speech (dots) encoder outputs (left), and shared encoder output (right)}
    \label{fig:shared}
    \vspace{-.3cm}
\end{figure}

\begin{figure}[!htb]
    \vspace{-.4cm}
    \centering
    \begin{minipage}{\columnwidth}
        \centering
        \includegraphics[width=.85\textwidth, height=5.5cm]{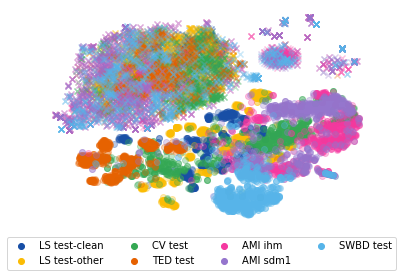}
    \end{minipage}%
    \vspace{-.25cm}
    \captionsetup{justification=centering}
    \caption{T-SNE of SLAM text (crosses) and speech (dots) shared encoder outputs }
    \label{fig:slam}
\end{figure}

To analyze representation spaces learned by Maestro and SLAM, we use a combination of visual inspection (via T-SNE~\cite{van2008visualizing}) and cosine similarity retrieval probes on mean-pooled speech and text embedding pairs. Although T-SNE is a low-dimension approximation of the overall high-dimension space, we see that the text and speech embeddings clearly appear to capture disparate regions, and different datasets reside in different areas of these regions (Figure \ref{fig:shared}). We see a unification of both modalities after projection through the shared encoder, and the sub-regions occupied by each dataset in those modalities are combined into one shared space. Compared to a baseline SLAM~\cite{bapna2021slam} text and speech encoding (Figure \ref{fig:slam}), the Maestro shared encoding indicates a larger unification of the two disparate information sources. While the included images are drawn from the LibriSpeech (LS) Maestro model, these effects are consistent across models for other domains.

\begin{figure}[!htb]
    \vspace{-.4cm}
    \centering
    \begin{minipage}{\columnwidth}
        \centering
        \includegraphics[width=.95\textwidth, height=4cm]{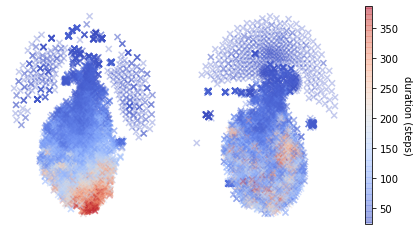}
    \end{minipage}%

    \vspace{-.25cm}
    \captionsetup{justification=centering}
    \caption{T-SNE of LS Maestro text encoder outputs (left), and shared encoder text output (right), color coded by duration}
    \label{fig:shareddur}
    \vspace{-.5cm}
\end{figure}

The combined joint space after Maestro's shared encoder is an ideal candidate for retrieval tests between speech and text examples to examine the learned representations, as is common in many other settings ~\cite{conneau2022fleurs,baevski2021unsupervised,bapna2021slam,bapna2022mslam,radford2021learning,hinton2011discovering}. We utilize candidate sets of $1,000$ paired speech and text examples from each dataset, testing for successful retrieval of the text which matches the speech example. These retrieval results are detailed in Table~\ref{t:srr}, and are used as a secondary inspection mechanism to understand the representation space. We leave larger considerations of retrieval as an explicit speech task for future work, focusing instead on simple exploratory retrieval probes to better understand the shared speech-text representations learned by Maestro and SLAM. 

Stronger retrieval performance indicates the effectiveness of the shared encoder to remove unwanted information and unify representations before passing to downstream speech tasks like ASR or Speech Translation. For example, we observe that duration information is strongly represented in the text encoder outputs (Figure~\ref{fig:shareddur}), but following the shared encoder, this influence is more diffuse.   We note that retrieval is generally a challenging inspection mechanism; non-unified spaces often have extremely degraded retrieval performance, rendering direct retrieval probes ineffective. For example, cosine retrieval using the baseline SLAM model from Figure~\ref{fig:slam} performs at or below $1$\% in the same settings as Table \ref{t:srr}. This makes the Maestro retrieval performance all the more remarkable. 

The effectiveness of this approach, along with related findings on multi-language retrieval~\cite{conneau2022fleurs}, opens a number of research questions about retrieval using modern multi-modal models, for example using shared spaces to directly sub-select vocabularies for ASR based on top-k retrieval results, or expanding the potential and limits of data augmentation in a shared space.
As an alternative to direct retrieval, ASIF~\cite{norelli2022asif} uses the concept of cross-modal anchoring in order to effectively retrieve between unimodal spaces. Using anchored retrieval allows probing the effectiveness of unimodal or non-unified spaces using paired multi-modal data, without further training. Higher ASIF retrieval indicates that two modal views of a data point are substantially related (relative to distractors) even if the respective feature spaces are not strictly unified.

This allows us to test the output of the speech and text encoders directly. $90\%$ of the target set act as anchors, while the remaining portion of the target set are used to evaluate. We repeat this procedure for $100$ trials of $100$ examples each ($10,000$ evaluations total), randomizing test points and anchors within the target set each trial and reporting the mean accuracy of the final result. 
Table~\ref{t:tsr} shows the results using $900$ anchors and $100$ retrievals per trial, with ASIF hyperparameters set to top k $90$ ($10$\%) and power $8$. We also report the accuracy of direct cosine retrieval under this setting, to highlight where anchoring is useful for non-unified spaces.

\begin{table}[!ht]
  \vspace{-.2cm}
  \caption{Shared space cosine retrieval probe accuracy (\%)}
  \label{t:shared_retrieval_results}
  \vspace{-.15cm}
  %vskip 0.1in
  \label{t:srr}
  \centering
  \small
  \resizebox{.975\columnwidth}{!}{%
  \begin{tabular}{lccccccc}
    \toprule
    \bfseries Method &
    \multicolumn{2}{c}{LibriSpeech} & \multicolumn{2}{c}{AMI} &
    \multicolumn{1}{c}{CV} & \multicolumn{1}{c}{SWBD} & \multicolumn{1}{c}{TED} \\
    \cmidrule{2-8}
    & test-clean & test-other & ihm & sdm1 & test & test & test\\
    \midrule
    \bfseries LS Maestro & 83.5 & 68.8 & 23.2 & 12.9 & 28.8 & 35.8 & 70.3 \\
    \bfseries AMI Maestro & 89.3 & 76.7 & \textbf{47.1} & \textbf{31.6} & 30.6 & 53.3 & 75.1 \\
    \bfseries CV Maestro & \textbf{97.5} & \textbf{96.5} & 32.1 & 15.8 & 72.4 & 48.6 & \textbf{94.9} \\
    \bfseries SWBD Maestro & 96.1 & 90.0 & 31.4 & 8.3 & 42.8 & \textbf{56.8} & 79.7 \\
    \bfseries TED Maestro & 95.8 & 94.5 & 38.2 & 19.3 & \textbf{76.0} & 53.2 & 91.7 \\
    \bfseries LS+C4 mSLAM & 0.30 & 0.20 & 0.00 & 0.30 & 0.10 & 0.00 & 0.30 \\
    \bottomrule
  \end{tabular}
  }
  \vspace{-.15cm}
\end{table}

\begin{table}[!ht]
  \vspace{-.2cm}
  \caption{Text and speech encoder retrieval probe accuracy (\%)}
  \vspace{-.15cm}
  \label{t:textspeech_retrieval_results}
  %vskip 0.1in
  \label{t:tsr}
  \centering
  \small
  \resizebox{.975\columnwidth}{!}{%
  \begin{tabular}{lccccccc}
    \toprule
    \bfseries Method &
    \multicolumn{2}{c}{LibriSpeech} & \multicolumn{2}{c}{AMI} &
    \multicolumn{1}{c}{CV} & \multicolumn{1}{c}{SWBD} & \multicolumn{1}{c}{TED} \\
    \cmidrule{2-8}
    & test-clean & test-other & ihm & sdm1 & test & test & test\\
    \midrule
    \bfseries LS Maestro (Direct) & 20.5 & 19.3 & 7.65 & 6.16 & 7.43 & 13.88 & 11.89 \\
    \bfseries LS Maestro (ASIF) & 45.7 & 31.2 & 7.47 & 5.61 & 10.2 & 10.76 & 16.64 \\
    \midrule
    \bfseries AMI Maestro (Direct) & 67.2 & 48.9 & 45.2 & 32.6 & 19.0 & 44.7 & 43.9 \\
    \bfseries AMI Maestro (ASIF) & 33.6 & 17.7 & 14.9 & 10.5 & 7.88 & 16.7 & 21.5 \\
    \midrule
    \bfseries CV Maestro (Direct) & 76.3 & 61.7 & 19.1 & 10.0 & 40.0 & 29.3 & 44.8 \\
    \bfseries CV Maestro (ASIF) & 50.4 & 34.8 & 14.3 & 7.61 & 20.1 & 19.4 & 28.9 \\
    \midrule
    \bfseries SWBD Maestro (Direct) & 20.3 & 14.1 & 15.9 & 8.61 & 10.3 & 25.3 & 13.8 \\
    \bfseries SWBD Maestro (ASIF) & 49.0 & 23.0 & 13.8 & 7.32 & 8.94 & 19.7 & 29.0 \\
    \midrule
    \bfseries TED Maestro (Direct) & 80.6 & 64.3 & 24.5 & 13.3 & 29.0 & 40.6 & 77.9 \\
    \bfseries TED Maestro (ASIF) & 43.6 & 26.9 & 13.3 & 7.96 & 11.8 & 16.8 & 25.8 \\
    \midrule
    \bfseries LS+C4 mSLAM (Direct) & 1.96 & 2.0 & 1.54 & 1.10 & 1.5 & 1.63 & 1.52 \\
    \bfseries LS+C4 mSLAM (ASIF) & 8.63 & 10.5 & 3.99 & 3.06 & 1.79 & 6.03 & 5.78 \\
    \bottomrule
  \end{tabular}
  }
  \vspace{-.15cm}
\end{table}

Here we find that relative representation retrieval (as in ASIF~\cite{norelli2022asif}) reveals the effectiveness of anchoring non-unified spaces particularly seeing improved performance when direct cosine retrieval is degraded (for example SWBD Maestro on TED in Table~\ref{t:tsr}). This also reinforces the visualization findings, indicating that the Maestro shared representation yields a substantially more unified representation space. The text and speech encoder representations, while related, remain distinct in many settings. While numbers across Table~\ref{t:srr} and Table~\ref{t:tsr} cannot be directly compared due to experimental differences, within table variations highlight the differences between direct cosine and ASIF retrieval approaches.
\vspace{-.3cm}

\section{Conclusion}
\vspace{-.2cm}
Through text-only domain adaptation and inspection of the learned speech and text representations, we draw the following conclusions and hypotheses.  Maestro enables effective speech-free domain adaptation across diverse corpora.
Our ablation studies demonstrate that duration/alignment are critical elements for successful adaptation. General purpose text encoders with corpus-specific duration models enable adaptation with minimal targeted customization. 
We hypothesize that the consistency MSE measure, text encoder loss, serves as an effective proxy for the effectiveness of a text-encoder for domain adaptation.
Furthermore, we inspect the representations learned by Maestro and SLAM across a variety of standard corpora, finding that modal encoders retain distinct information for each modality: speech and text. However, after shared encoding the Maestro modal representations are unified, and the overall shared space is a powerful, generic representation of speech or text, based on both visual inspection and retrieval probe tasks. We hypothesize that the coherence of this output space in joint speech-text representation learning in Maestro enables its power for text-only adaptation and state-of-the-art performance in both automatic speech recognition (ASR) and speech translation (ST).

\bibliographystyle{IEEEbib}
\bibliography{refs}

\begin{thebibliography}{10}

\bibitem{bapna2021slam}
Ankur Bapna, Yu-an Chung, Nan Wu, Anmol Gulati, Ye~Jia, Jonathan~H Clark,
  Melvin Johnson, Jason Riesa, Alexis Conneau, and Yu~Zhang,
\newblock ``{SLAM}: A unified encoder for speech and language modeling via
  speech-text joint pre-training,''
\newblock {\em arXiv preprint arXiv:2110.10329}, 2021.

\bibitem{bapna2022mslam}
Ankur Bapna, Colin Cherry, Yu~Zhang, Ye~Jia, Melvin Johnson, Yong Cheng, Simran
  Khanuja, Jason Riesa, and Alexis Conneau,
\newblock ``{mSLAM}: Massively multilingual joint pre-training for speech and
  text,''
\newblock {\em arXiv preprint arXiv:2202.01374}, 2022.

\bibitem{chen2022maestro}
Zhehuai Chen, Yu~Zhang, Andrew Rosenberg, Bhuvana Ramabhadran, Pedro Moreno,
  Ankur Bapna, and Heiga Zen,
\newblock ``Maestro: Matched speech text representations through modality
  matching,''
\newblock {\em arXiv preprint arXiv:2204.03409}, 2022.

\bibitem{sainath2022joist}
Tara~N. Sainath, Rohit Prabhavalkar, Ankur Bapna, Yu~Zhang, Zhouyuan Huo,
  Zhehuai Chen, Bo~Li, Weiran Wang, and Trevor Strohman,
\newblock ``Joist: A joint speech and text streaming model for asr,''
\newblock in {\em IEEE SLT}, 2022.

\bibitem{meng2022mhat}
Zhong Meng, Tongzhou Chen, Rohit Prabhavalkar, Yu~Zhang, Gary Wang, Kartik
  Audhkhasi, Jesse Emond, Trevor Strohman, Bhuvana Ramabhadran, W.~Ronny Huang,
  Ehsan Variani, Yinghui Huang, and Pedro~J. Moreno,
\newblock ``Modular hybrid autoregressive transducer,''
\newblock in {\em to appear in IEEE SLT}, 2022.

\bibitem{meng2022ilma}
Zhong Meng, Yashesh Gaur, Naoyuki Kanda, Jinyu Li, Xie Chen, Yu~Wu, and Yifan
  Gong,
\newblock ``Internal language model adaptation with text-only data for
  end-to-end speech recognition,''
\newblock in {\em Interspeech}, 2022.

\bibitem{thomas2022textogram}
Samuel Thomas, Brian Kingsbury, George Saon, and Hong-Kwang~J. Kuo,
\newblock ``Integrating text inputs for training and adapting rnn transducer
  asr models,''
\newblock in {\em IEEE ICASSP}, 2022.

\bibitem{udagawa2022rescoring}
Takuma Udagawa, Masayuki Suzuki, Gakuto Kurata, Nobuyasu Itoh, and George Saon,
\newblock ``Effect and analysis of large-scale language model rescoring on
  competitive asr systems,''
\newblock in {\em Interspeech}, 2022.

\bibitem{graves2012sequence}
Alex Graves,
\newblock ``Sequence transduction with recurrent neural networks,''
\newblock {\em arXiv preprint arXiv:1211.3711}, 2012.

\bibitem{graves2013speech}
Alex Graves, Abdel-rahman Mohamed, and Geoffrey Hinton,
\newblock ``Speech recognition with deep recurrent neural networks,''
\newblock in {\em 2013 IEEE international conference on acoustics, speech and
  signal processing}. IEEE, 2013, pp. 6645--6649.

\bibitem{chen2022maestro-u}
Zhehuai Chen, Ankur Bapna, Andrew Rosenberg, Yu~Zhang, Bhuvana Ramabhadran,
  Pedro Moreno, and Nanxin Chen,
\newblock ``Maestro-u: Leveraging joint speech-text representation learning for
  zero supervised speech asr,''
\newblock in {\em IEEE SLT}, 2022.

\bibitem{chan2021speechstew}
William Chan et~al.,
\newblock ``{SpeechStew}: Simply mix all available speech recognition data to
  train one large neural network,''
\newblock {\em arXiv preprint arXiv:2104.02133}, 2021.

\bibitem{sato2022textonly}
Hiroaki Sato, Tomoyasu Komori, Takeshi Mishima, Yoshihiko Kawai, Takahiro
  Mochizuki, Shoei Sato, and Tetsuji Ogawa,
\newblock ``Text-only domain adaptation based on intermediate ctc,''
\newblock in {\em Interspeech}, 2022.

\bibitem{chuang2019speechbert}
Yung-Sung Chuang, Chi-Liang Liu, Hung-Yi Lee, and Lin-shan Lee,
\newblock ``Speechbert: An audio-and-text jointly learned language model for
  end-to-end spoken question answering,''
\newblock {\em arXiv preprint arXiv:1910.11559}, 2019.

\bibitem{sunder2022tokenwise}
Vishal Sunder, Eric Fosler-Lussier, Samuel Thomas, Hong-Kwang~J. Kuo, and Brian
  Kingsbury,
\newblock ``Tokenwise contrastive pretraining for finer speech-to-bert
  alignment in end-to-end speech-to-intent systems,''
\newblock in {\em Interspeech}, 2022.

\bibitem{hotelling1936relations}
H.~Hotelling,
\newblock ``Relations between two sets of variates,''
\newblock {\em Biometrika}, vol. 28(3/4), pp. 321–377, 1936.

\bibitem{sun2020dcca}
Zhongkai Sun, Prathusha~Kameswara Sarma, William~A. Sethares, and Yingyu Liang,
\newblock ``Learning relationships between text, audio, and video via deep
  canonical correlation for multimodal language analysis,''
\newblock in {\em The Thirty-Fourth {AAAI} Conference on Artificial
  Intelligence, {AAAI} 2020, The Thirty-Second Innovative Applications of
  Artificial Intelligence Conference, {IAAI} 2020, The Tenth {AAAI} Symposium
  on Educational Advances in Artificial Intelligence, {EAAI} 2020, New York,
  NY, USA, February 7-12, 2020}. 2020, pp. 8992--8999, {AAAI} Press.

\bibitem{hu2020multilingual}
Yushi Hu, Shane Settle, and Karen Livescu,
\newblock ``Multilingual jointly trained acoustic and written word
  embeddings,''
\newblock in {\em Interspeech}, 2020.

\bibitem{kamper2016acousticword}
Herman Kamper, Weiran Wang, and Karen Livescu,
\newblock ``Deep convolutional acoustic word embeddings using word-pair side
  information,''
\newblock in {\em ICASSP}, 2016.

\bibitem{zheng2021fused}
Renjie Zheng et~al.,
\newblock ``Fused acoustic and text encoding for multimodal bilingual
  pretraining and speech translation,''
\newblock {\em arXiv preprint arXiv:2102.05766}, 2021.

\bibitem{kahn2020libri}
Jacob Kahn, Morgane Rivi{\`e}re, Weiyi Zheng, et~al.,
\newblock ``{Libri-Light}: A benchmark for asr with limited or no
  supervision,''
\newblock in {\em Proc. ICASSP}, 2020, pp. 7669--7673.

\bibitem{chung2021w2v}
Yu-An Chung et~al.,
\newblock ``{W2v-BERT}: Combining contrastive learning and masked language
  modeling for self-supervised speech pre-training,''
\newblock {\em arXiv preprint arXiv:2108.06209}, 2021.

\bibitem{van2008visualizing}
Laurens Van~der Maaten and Geoffrey Hinton,
\newblock ``Visualizing data using t-sne.,''
\newblock {\em Journal of machine learning research}, vol. 9, no. 11, 2008.

\bibitem{conneau2022fleurs}
Alexis Conneau, Min Ma, Simran Khanuja, Yu~Zhang, Vera Axelrod, Siddharth
  Dalmia, Jason Riesa, Clara Rivera, and Ankur Bapna,
\newblock ``Fleurs: Few-shot learning evaluation of universal representations
  of speech,''
\newblock {\em arXiv preprint arXiv:2205.12446}, 2022.

\bibitem{baevski2021unsupervised}
Alexei Baevski, Wei-Ning Hsu, Alexis Conneau, and Michael Auli,
\newblock ``Unsupervised speech recognition,''
\newblock {\em Advances in Neural Information Processing Systems}, vol. 34, pp.
  27826--27839, 2021.

\bibitem{radford2021learning}
Alec Radford, Jong~Wook Kim, Chris Hallacy, Aditya Ramesh, Gabriel Goh,
  Sandhini Agarwal, Girish Sastry, Amanda Askell, Pamela Mishkin, Jack Clark,
  et~al.,
\newblock ``Learning transferable visual models from natural language
  supervision,''
\newblock in {\em International Conference on Machine Learning}. PMLR, 2021,
  pp. 8748--8763.

\bibitem{hinton2011discovering}
Geoffrey Hinton and Ruslan Salakhutdinov,
\newblock ``Discovering binary codes for documents by learning deep generative
  models,''
\newblock {\em Topics in Cognitive Science}, vol. 3, no. 1, pp. 74--91, 2011.

\bibitem{norelli2022asif}
Antonio Norelli, Marco Fumero, Valentino Maiorca, Luca Moschella, Emanuele
  Rodol{\`a}, and Francesco Locatello,
\newblock ``Asif: Coupled data turns unimodal models to multimodal without
  training,''
\newblock {\em arXiv preprint arXiv:2210.01738}, 2022.

\end{thebibliography}

\end{document}